\documentclass[11pt,a4paper]{article}
\usepackage{times,latexsym}
\usepackage{url}
\usepackage[T1]{fontenc}
\usepackage{multirow, amsmath, booktabs, paralist}
\usepackage{float}
\usepackage{longtable} 

\usepackage{xspace,mfirstuc,tabulary}
\usepackage{times}
\usepackage{latexsym}
\usepackage[T1]{fontenc}
\usepackage[utf8]{inputenc}
\usepackage{microtype}
\usepackage{inconsolata}
\usepackage{graphicx}
\usepackage{hyperref}
\usepackage{cleveref}
\usepackage{subcaption}
\usepackage{stfloats} 
\usepackage[acceptedWithA]{tacl2021v1}

\newif\iftaclinstructions
\taclinstructionsfalse 
\iftaclinstructions
\renewcommand{\confidential}{}
\renewcommand{\anonsubtext}{(No author info supplied here, for consistency with
TACL-submission anonymization requirements)}
\newcommand{\instr}
\fi

\iftaclpubformat 

\else

\fi

\title{Retain or Reframe? A Computational Framework for the Analysis of Framing in News Articles and Reader Comments}

\author{
  Matteo Guida \quad
  Yulia Otmakhova \quad
  Eduard Hovy \quad
  Lea Frermann
  \\
  School of Computing and Information Systems,\\
  The University of Melbourne\\
  \texttt{m.guida@student.unimelb.edu.au} \\
  \texttt{\{y.otmakhova, eduard.hovy, lea.frermann\}@unimelb.edu.au}
}

\date{}

\begin{document}
\maketitle

\begin{abstract}
When a news article describes immigration as an "economic burden" or a "humanitarian crisis," it selectively emphasizes certain aspects of the issue. Although this \textit{framing} shapes how the public interprets such issues, audiences do not absorb frames passively but actively reorganize the presented information. While this relationship between source content and audience response is well-documented in the social sciences, NLP approaches often ignore it, detecting frames in articles and responses in isolation. 
We present the first computational framework for large-scale analysis of framing across source content (news articles) and audience responses (reader comments). Methodologically, we refine frame labels and develop a framework that reconstructs dominant frames in articles and comments from sentence-level predictions, and aligns articles with topically relevant comments.
Applying our framework across eleven topics and two news outlets, we find that frame reuse in comments correlates highly across outlets, and that readers often selectively engage with frames of the articles. We release a frame classifier that performs well on both articles and comments, a dataset of article and comment sentences manually labeled for frames, and a large-scale dataset of articles and comments with predicted frame labels.\footnote{We released our data at: \url{https://github.com/mattguida/FrAC}; and fine-tuned frame classifier: \url{https://huggingface.co/mattdr/sentence-frame-classifier}}

\end{abstract}

\section{Introduction}
Framing research has long recognized that how news content is presented fundamentally shapes public understanding of issues \citep{entman1993framing, chong2007framing}. However, audiences do not passively absorb frames but rather "\textit{actively filter, sort and reorganize information in personally meaningful ways in the process of constructing an understanding of public issues}" \citep[pg.77]{neuman1992common}. {The constructionist understanding of framing emphasizes that frames cannot be fully understood without considering the interaction between source and receiver \citep{scheufele1999framing, vangorp2007constructionist, hubner2023news}, as part of an interactive and dynamic process in which frames activate and reasonate (or fail to) with personal schemas through active meaning construction and intepretation. This is particularly evident on news websites with comment threads that create immediate opportunities for readers to publicly respond to framed content -- a paradigm shift from one-way communication to participatory discourse where audiences actively engage with and reconstruct news content~\citep{risch_dataset_2020}. 

However, current research has yet to fully explore the relationship between media framing and audience reception. Cognitive studies have demonstrated that the framing of news articles can significantly shape readers' perceptions of an issue in controlled, small-scale experiments \citep{price1997switching, valkenburg1999, nelson1999issue}. Communication scholars have examined the influence of framing by analyzing online news articles and their associated reader comments \citep{zhou_parsing_2007, muniz_shaping_2015, dargay2016, hubner2023news}. These studies often focused on a single issue and relied on small, manually annotated data sets, which limited generalizability and led to inconsistent findings.

Meanwhile, Natural Language Processing (NLP) approaches have developed numerous classification systems (for a review, see \citet{ali_survey_2022}) aimed at detecting frames in text of different lengths such as news articles \citep{field_framing_2018}, or social media posts \citep{mendelsohn_modeling_2021}. Yet these large-scale methods treat frame detection in source documents (such as news) and responses to them (such as comments or social media posts) as isolated classification tasks without connecting the framed source text and the framing in the audience response \citep{otmakhova_frermann_media_2024}.  

\begin{figure}
    \centering
    \includegraphics[width=\linewidth,clip,trim=0.0cm 0 0 0cm]{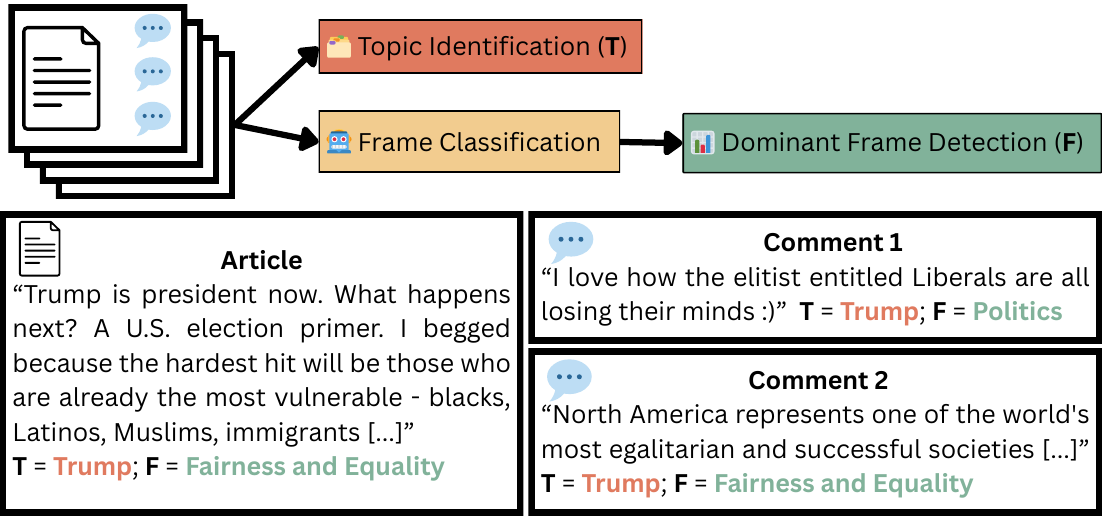}
    \caption{Overview of our framework. Bottom: Article with two comments illustrating {\bf reframing} (Comment~1; changes the article frame) and {\bf frame retention} (Comment~2; keeps the article frame).}
    \label{fig:framework}
\end{figure}

{If framing is truly a process in which frames are actively constructed and interpreted by an audience \citep{scheufele1999framing, vangorp2007constructionist, hubner2023news}, then the audience's response is not merely a posited outcome, but a fundamental part of the analysis that has been largely overlooked or observed at a limited scale. To address this gap, we propose a computational framework that investigates at scale the relationship between how information is framed in news articles and how readers respond to such framed content in associated comments. Using an established framing taxonomy \citep{boydstun_tracking_2014, card_media_2015}, we examine both source content and audience responses as interconnected elements of a broader communicative process \citep{hubner2023news}.

{Throughout this paper, we use the phrase \textbf{media framing effects} to describe the relation between news article framing and that of reader comments, as an umbrella term that covers two phenomena. \textbf{Frame retention} occurs when readers adopt the dominant frame\footnote{We define the dominant frame as the one that covers the most sentences of an article, but at least 40\% (see Section~\ref{sec:dominant}).} used in the source article in their comments. \textbf{Reframing} occurs when readers shift to a different frame in their comments: either a minor frame in the article ({\textit{selective} reframing}), or a completely new, unmentioned frame ({\textit{complete} reframing}). Figure~\ref{fig:framework} illustrates this with an example.

Our large-scale analysis across eleven diverse topics and two news outlets (The New York Times and The Globe and Mail) reveals systematic patterns of media framing effects. We find that 
(a) frame retention rates vary by source document frame, and this variation correlates highly across the two outlets, with value-laden frames being more often reframed; (b) retention patterns vary systematically by  topic, and (c) audiences engage in selective reframing, adopting secondary frames rather than completely accepting or rejecting primary frames.

We make three main contributions: (1) a computational framework for investigating the relationship between source frames (articles) and frames in associated responses (comments), (2) a frame classification model that effectively generalizes across different text types (news articles and reader comments) and topics, and (3) two data sets: a manually labeled evaluation corpus (FrAC) and a large-scale data set of articles and comments aligned by topic group and labeled with dominant frames, for reproducibility and supporting future research.

\section{Related Work}
\label{sec:related_work}
\paragraph{Cognitive Studies on Framing Effects} Cognitive research has established that news frames significantly influence audience interpretation using controlled experimental designs, where participants were assigned to different framing conditions of the same issue and asked to report their thoughts. \citet{valkenburg1999} examined how news stories framed as conflict, human interest, attribution of responsibility, or economic consequences affected readers, finding that participants generally adopted news frames. However, \citet{price1997switching} showed that frames not only guide interpretation but also stimulate new thoughts and feelings not explicitly present in the original content. \citet{nelson1999issue} and \citet{shen2004effects} showed that framing primarily influences the importance of existing beliefs rather than changing them, with frames being the most effective when they resonate with pre-existing schemas. These studies were restricted to small-scale experimental settings and narrow demographic representation, and thus could not generalize to natural media environments with direct audience responses.

\paragraph{Media Framing Effects in Communication Studies}
\label{sec:comm_studies}
Communication researchers have attempted to bridge this gap by analyzing reader comments on online news articles, albeit with mixed results.
Several studies have found evidence of frame retention. \citet{kleut_framing_2021} analyzed online coverage of Serbian protests and showed that comments were influenced by the article's frames.  \citet{dargay2016} examined terrorism coverage in The New York Times, finding that commentators engaged with dominant frames of the articles, although they also introduced new frames.

Other studies have found no evidence of retention. \citet{holton_commenting_2014} analyzed online health coverage across major outlets and found no evidence of frame retention in comment threads. \citet{muniz_shaping_2015} examined coverage of an oil company expropriation in Spanish online outlets and found that media frames did not produce corresponding effects in reader comments. Similarly, \citet{zhou_parsing_2007} found no framing effects in news about Chinese sociopolitical incidents. Finally, in climate change coverage, \citet{koteyko_climate_2013} found minimal retention of frames in reader comments on online articles from UK outlets, while \citet{hubner2023news} showed that retweets sharing climate news rarely replicated the original frames. 

There are several explanations for these inconsistencies, including small data sets, focus on a single topic, and the different adopted framing conceptualizations. In this study, we increase the data set size, keep the framing conceptualization fixed but apply it across eleven topics, in order to disentangle some of these confounds.

\paragraph{Computational Approaches}
NLP approaches to framing have evolved from exploratory topic-modeling \citep{DiMaggio2013ExploitingAB} to classification tasks (for a review, see \citet{ali_survey_2022}). The Media Frame Corpus (MFC, \citet{card_media_2015}), based on \citet{boydstun_tracking_2014}, has been particularly influential, spurring various frame classification methods including probabilistic approaches (e.g.,\citet{johnson_classification_2018}), neural networks (e.g., \citet{naderi2017classifying}), and pre-trained language models (e.g., \citet{khanehzar_framing_2021}). These computational approaches have analyzed frames at different granularities: headline level \citep{liu2019detecting, akyurek2020multi}, document level \citep{field_framing_2018, ji2017neuraldiscoursestructuretext, piskorski_multilingual_2023}, span level \citep{khanehzar_framing_2021}, sentence level \citep{naderi2017classifying, hartmann_issue_2019}, and paragraph level \citep{roy-goldwasser-2020-weakly}. 
Some studies have investigated the generalizability of the MFC to other domains and languages: \citet{khanehzar_modeling_2019} found reduced performance when applying fine-tuned BERT-based classifiers to different English contexts (Australian parliamentary speeches), while \citet{daffara2025generalizabilitymediaframescorpus} tested cross-linguistic transfer using both fine-tuned pre-trained language models and zero-shot LLM classification on Portuguese news articles. However, these studies either focused on cross-linguistic transfer or on the \textit{diagnosis} of the problem of cross-topic and genre generalization. They fell short of offering a solution to the problem, which we do here.

\section{Research Questions}
\label{sec:rqs}
Our goal is to characterize and quantify the {\it constructivist view on framing} at scale. We formulate research questions that have been raised in the social sciences, albeit on a small scale often regarding a single topic and a specific framing inventory which makes it difficult to generalize results across scenarios. With orders of magnitude more data that covers a range of topics and is analyzed with a unified framing inventory, we are well positioned to answer the following research questions.
{\begin{center}
    {\it RQ1: To what extent are dominant article frames retained in reader comments?}
\end{center}
We first focus on the extent to which the most prominent frame in the article is reflected in the reader comments. From a constructivist point of view, this assesses the extent to which a chosen article frame {\it sticks with} the readership.

Next, we address the fact that prior work was often topic-specific and led to inconclusive results about the extent of frame retention. Here, we can quantify the effect of topics on frame retention within a unified framework. We ask
\begin{center}
    {\it RQ2: How does frame retention vary across topics?}
\end{center}

Finally, we examine {\it how} readers how audiences reconstruct media messages when they do reframe (i.e., they choose not to retain the dominant article frame). Our framework allows us to paint a more nuanced picture on reframing than previous work. First, in line with prior studies we quantify how often readers adopt a frame that was never used in the article ({\it complete reframing}). In addition, we also explore how often readers shift from the dominant article frame to a {secondary article frame} ({\it selective reframing}). This leads to our final research question:

\begin{center}
    {\it RQ3: If readers reject the dominant article frame, how do they reframe?}
\end{center}

We next introduce our framework and methodology, before we address our research question in Section~\ref{sec:analysis}.

\section{Framework} 
\label{sec:framework} 
To address these research questions, we propose a scalable and automated framework for analyzing media framing effects (retention and reframing). 


\subsection{Data}
\label{ssec:data}
Our framework requires a data set of source documents $d_i$ (e.g., news articles, headlines, tweets, etc.) each paired with a set of response documents $R_{d_i} = \{r_{1}, r_{2}, \ldots\}$ (e.g., comments, retweets, blog posts, etc.) directly linked to the source ($D = \{(d_i, R_{d_i})\}_{i=1}^{N}$). For the purpose of this study, we use two existing data sets of news articles with their associated comments: 

\begin{compactitem}
    \item \textbf{SOCC} (The SFU Opinion Comments Corpus; \citet{kolhatkar2020sfu})\footnote{Available at: \url{https://www.kaggle.com/datasets/mtaboada/sfu-opinion-and-comments-corpus-socc}}: consists of 10,339 articles and over 1.2 million comments sourced from The Globe and Mail, a major Canadian newspaper. The corpus covers five years (2012–2016).
    
     \item \textbf{NYT} (New York Times): consists of articles and comments from \textit{The New York Times}\footnote{Available at: \url{https://www.kaggle.com/datasets/aashita/nyt-comments}. Article texts were retrieved using the article titles in the data set via ProQuest.} with over two million comments associated with approximately 9,000 articles published between January-May 2017 and January-April 2018.
\end{compactitem}
This pairing of source documents with response documents directly operationalizes the constructionist understanding of framing. In our analysis, news articles represent the 'source framing' -- how journalists and media outlets choose to present issues through particular frames. Reader comments represent the 'audience response' -- how readers actively interpret, accept, modify, or reject those frames based on their own cognitive schemas and values. We focus exclusively on top-level comments that directly respond to articles, excluding nested replies within comment threads, as we want to capture direct audience responses to the original framed content.

\subsection{Framework Structure}

The framework consists of three core components that work in concert to analyze framing dynamics between articles and comments (\Cref{fig:framework}):

\begin{compactenum}    
    \item \textbf{Frame Classification}: To compare framing across source documents and responses, we build on established framing taxonomies to develop and evaluate frame classifiers that work reliably across both document types.
    
    \item \textbf{Dominant Frame Detection}: In line with prior work in cognitive and communication studies (Section~\ref{sec:related_work}), we identify and compare a single, {\it dominant} frame (defined as the most recurring one) in source documents and responses. We thus require a method to determine the dominant frame in each input document. We define the dominant frame as the most frequently occurring sentence-level frame within a document (see Section~\ref{sec:dominant} for details). In doing so, we can distinguish the retention of the primary article frame from {\it selective reframing} into a {\it secondary frame} of the original article.
    \item \textbf{Topic Identification}: 
    We assign each source document and reader response to a topic, for two reasons.
    First, assigning topics to articles allows us to perform analysis of framing effects over multiple topics, advancing over previous, topic-specific studies (\Cref{sec:comm_studies}) and potentially explaining their divergent results. Second, we want to capture framing effects only for responses where commentators stay on topic with the source, rather than digress into a different topic (with potentially a different frame, which can lead to underestimation of retention rates).

\end{compactenum} 
Below, we first introduce a new corpus that we annotated for the \textit{frame classification} subtask (\Cref{sec:frac}). \Cref{sec:methodology} explains and evaluates our implementations of the three framework components introduced above.
Finally, we apply the framework to construct a data set (\Cref{sec:result_dataset}) that we use to analyze media framing effects (\Cref{sec:analysis}).


\section{The FrAC Corpus}
\label{sec:frac}

To apply our framework, we require (1) a frame label set that works robustly across both source documents (news articles) and reader reactions (comments), and across a variety of different topics; and (2) a way to validate predicted frames across these two types of documents.  To this end, we introduce the {\bf Fr}ames in {\bf A}rticles and {\bf C}omments (FrAC) corpus. Our corpus uses a more robust frame label set derived from that of the Media Frames Corpus (Section~\ref{ssec:label}), and has high-quality manual annotations showing that humans can reliably apply our label set to both document types (Section~\ref{ssec:annotation}).



\subsection{Frame Labels and Granularity}
\label{ssec:label}
We devise a robust frame label set, building on the Media Frames Corpus (MFC), a foundational data set in computational framing analysis based on the \citet{boydstun_tracking_2014}'s theoretical framework operationalizing framing through 15 generic frame categories. The original MFC corpus~\cite{card_media_2015} consists only of news articles; however, more recently the framework has been successfully applied to other types of documents, including online forum comments~\cite{hartmann_issue_2019}, suggesting suitability for our task that involves both long, formal documents and short, less formal responses.


First, to apply the MFC labels across articles and comments, we need to define a granularity that is meaningful for both of them. While most NLP studies on the MFC modeled article-level dominant frames, this is not feasible for reader responses which are often brief and composed of single sentences. We therefore chose the sentence level -- the smallest shared textual unit between the two types of text -- as our basic unit of analysis.\footnote{\Cref{sec:dominant} explains how we derive a dominant document frame from sentence-level frames.}

Second, we improve the robustness of the MFC label set as it has been found (e.g., \citet{ali_survey_2022, daffara2025generalizabilitymediaframescorpus}) that some MFC frames are poorly defined and difficult to distinguish, leading to low inter-annotator agreement and model performance. 

To do so, we isolated highly confused frame labels in the original MFC data set by first identifying span-level MFC annotations (of ${\geq}3$ words) to which two or more annotators assigned different frame labels. We examined such inconsistent annotations using confusion matrices -- both for all MFC data and separately for each topic included in the MFC corpus. We took a conservative approach by only merging frame labels that were confused consistently across all topics,\footnote{For example, ``Health and Safety'' and ``Legality, constitutionality and jurisprudence'' labels are frequently assigned to the same spans in ``Tobacco'' and ``Gun Control'', but not in other topics. In this case, we conclude that the confusion is due to the topic specifics rather than meaning of the label, and keep such semantically distinct labels separate.}  resulting in two merges of two frames (\Cref{tab:frames}).
To further improve robustness of the label set, we removed the two least frequent classes (``External regulation and reputation'' and ``Capacity and resources'') which accounted for less than 0.7\% of the data. 
We finally obtained 9 frame categories, plus a generic ``Other'' frame to account for instances that do not match any of our nine frames. This category captures both genuinely unframed content and framed content that falls outside our nine established categories.  
Table~\ref{tab:frames} lists the frames, with the definitions following those of \ref{tab:framing_categories}, as depicted in the Appendix by Table~\ref{tab:framing_categories}.

\begin{table}[]
    \centering
    \small
    \begin{tabular}{|p{0.95\columnwidth}|}
    \toprule
    1) Legality and Crime$^*$, 2) Political and Policies$^+$, \\
    3) Economic, 4) Health and Safety, 5) Cultural Identity, 6) Public Opinion, 7) Morality, 8) Fairness and Equality, 9) Security and Defense, 10) Other. \\
    \bottomrule
    \end{tabular}
    \caption{The FrAC frame label set. Label mergers: $* {=} \{$"Legality, constitutionality and jurisprudence", "Crime and punishment"$\}$; $+{=}\{$"Policy prescription and evaluation", "Political"$\}$.}
    \label{tab:frames}
\end{table}

\subsection{Human Annotation}
\label{ssec:annotation}
To validate our frame label set (and model predictions) we collected human annotation on a subset of the SOCC corpus (\Cref{ssec:data}). 

\paragraph{Annotation Setup} 
A custom web-based annotation interface was developed using FastAPI.\footnote{\url{https://annotation-app-slty.onrender.com/}.} The interface included detailed instructions and examples, and an interactive training phase with immediate feedback and explanations. See Appendix~\ref{annotation-interface} for detailed descriptions.

We recruited six native English speakers as annotators, all with a background in NLP or political science and journalism.\footnote{This study was approved by Human Ethics Committee (Referance No. 2025-32561-67793-4)
and has been taken out to the according ethical standards. Annotators were compensated with vouchers over 50 USD, well above local minimum wage requirements.}  
Each annotator was presented with 100 sentences from the SOCC corpus, drawn evenly from articles and comments, 20\% of the sentences overlapped with another annotator to assess agreement. One additional annotator (an author of this paper) annotated all sentences (so all sentences in the resulting data set had at least two annotations). We adopted a multi-label annotation setup. Annotators were presented one input sentence at a time, and were asked to select at least one out of our 9 frames. If no frame matched, the category ``Other'' should be selected.

\paragraph{Inter-Annotator Agreement} Given our multi-label setup, we assessed inter-annotator agreement in two ways. Following prior work~\cite{arora_multi-modal_2025}, we use label-wise Krippendorff's $\alpha$ to assess per-label agreement, and the Jaccard index to measure the averaged overlap in label sets between annotators. The averaged Krippendorff's $\alpha$ across all labels and annotators indicates moderate to high agreement ($\alpha = 0.76$), with some variance across frame labels (Table~\ref{tab:krippendorff_per_label_sorted}), however even the frame category with the lowest $\alpha$ reaches moderate agreement. 
We also compute an averaged Jaccard index of 0.79 which confirms this trend, suggesting a substantial overlap across label sets. This shows that a more robust set of labels applied at a sentence level allows to achieve substantial agreement despite the purported subjectivity of the task.

\begin{table}[t]
\centering
\small
\setlength{\tabcolsep}{3pt}
\begin{tabular}{lc|lc}
\toprule
\textbf{Label} & \textbf{$\alpha$} &\textbf{Label} & \textbf{$\alpha$}\\
\midrule
{Economic}             & 0.89 &{Political / Policy} & 0.78 \\
{Health / Safety}    & 0.87 &{Cultural Identity}    & 0.76 \\
{Legality / Crime}   & 0.82 &{Other}                & 0.66 \\
{Security / Defense} & 0.79 &{Fairness / Equality}& 0.65 \\
{Public Opinion}       & 0.79 &{Morality}             & 0.63 \\
\bottomrule
\end{tabular}
\caption{Krippendorff's $\alpha$ per label, averaged across annotation sessions and sorted in descending order.}
\label{tab:krippendorff_per_label_sorted}
\end{table}


\paragraph{Gold Standard} To construct a gold standard for model evaluation, we applied majority voting among annotators for each sentence. 
A clear consensus (where either two or three annotators chose the same label) was achieved for 532 cases (where in 500 cases a single label was assigned, while in 32 all annotators have chosen two consistent labels). Sentences where annotators disagreed (45) were adjudicated by two authors of the paper, who independently re-assessed the cases of contention, choosing the label out of 9 categories. We kept only the cases where both adjudicators agreed with one of the original annotations, discarding 13 sentences. This process resulted in a validation data set of 556 sentences, which we use in \Cref{sec:mfc_theory_classification} to assess model performance.


\section{Methodology}
\label{sec:methodology}
Having established a general framework, and a conceptualization of media frames that generalize across articles and comments, we next describe the implementation of the framework components on our SOCC and NYT data sets of articles and comments (Section \ref{ssec:data}).


\subsection{Frame Classification}
\label{sec:mfc_theory_classification}

We train frame classifiers that predict sentence-level media frames for both comments and articles, following the framing schema defined in Section~\ref{ssec:label}. To do so, we require suitable training data from both document types, and a robust evaluation setup.

\paragraph{Training Data}
As our FrAC corpus is too small to support fine-tuning, we derive large-scale labeled data sets from the MFC (news articles) and \citet{hartmann_issue_2019} (comments).
We sentence-split both datasets and derive sentence-level annotations from the original span-level annotations: we first map the original frame labels to our 9 classes; then we identify spans with matching labels from at least two annotators with a minimum three-word overlap. Following previous methods \citep{naderi2017classifying, hartmann_issue_2019, park_challenges_2022}, agreed spans were mapped to the sentence-level: sentences fully covered by agreed spans inherited the span labels. This results in approximately 63,000 labeled sentences predominantly from the MFC (including 625 sentences  from the much smaller \citet{hartmann_issue_2019} dataset). Further details on frame and topic distributions of the training data are provided in Appendix~\ref{app:processed_training}, Table~\ref{tab:frame_distribution}.

\paragraph{Frame Classification Models}
To automatically classify frames in news articles and comments, we fine-tuned three models -- \texttt{RoBERTa-Large} \citep{liu2019robertarobustlyoptimizedbert} (full fine-tuning), \texttt{LLaMA3.1-8B-Instruct} \citep{grattafiori2024llama}, and \texttt{Qwen2.5-7B-Instruct} \citep{qwen2025qwen25technicalreport} (both parameter-efficient fine-tuning) -- using the sentence-level annotated data described above with a random 80/10/10 (training/dev/test) split, stratified by frame and topic. 
Fine-tuning parameters are reported in Appendix~\ref{app:fine-tune}, and prompting details in Appendix~\ref{app:prompt}.

\begin{table}[t] 
\centering 
\setlength{\tabcolsep}{3pt}
\footnotesize 
\begin{tabular}{ccccccccccc} 
\toprule 
\multicolumn{3}{c}{\textbf{RoBERTa}} && \multicolumn{3}{c}{\textbf{Llama}} && \multicolumn{3}{c}{\textbf{Qwen}}\\ 
\textbf{P}&\textbf{R}&\textbf{F1}&&\textbf{P}&\textbf{R}&\textbf{F1}&&\textbf{P}&\textbf{R}&\textbf{F1}\\
\midrule 
 0.70 & 0.65 & \textbf{0.66} && 0.68 & 0.65 & {0.65} && 0.64 & 0.64 & {0.64} \\
\bottomrule 
\end{tabular}
\caption{Precision (P), recall (R) and Macro-F1 scores for the three fine-tuned LMs on MFC + Hartmann data.}
\label{tab:f1_avg} 
\end{table}


\begin{table}[t]
\centering
\small
\setlength{\tabcolsep}{6pt} 
\begin{tabular}{lccc}
\toprule
\textbf{Label} & \textbf{P} & \textbf{R} & \textbf{F1} \\
\midrule
\textbf{Economic}            & 0.77 & 0.83 & 0.80 \\
\textbf{Morality}            & 0.70 & 0.70 & 0.70 \\
\textbf{Fairness and Equality} & 0.68 & 0.59 & 0.64 \\
\textbf{Legality and Crime}  & 0.83 & 0.86 & 0.85 \\
\textbf{Political and Policy} & 0.77 & 0.85 & 0.81 \\
\textbf{Security and Defense} & 0.69 & 0.59 & 0.63 \\
\textbf{Health and Safety}   & 0.73 & 0.80 & 0.76 \\
\textbf{Cultural Identity}   & 0.71 & 0.54 & 0.61 \\
\textbf{Public Opinion}      & 0.61 & 0.61 & 0.61 \\
\textbf{Other}               & 0.53 & 0.14 & 0.22 \\
\bottomrule
\end{tabular}
\caption{Per-frame precision, recall, and Macro-F1 scores for fine-tuned RoBERTa on MFC + Hartmann data.}
\label{tab:f1_all}
\end{table}

\begin{table}[t] 
\centering 
\setlength{\tabcolsep}{3pt}
\footnotesize 
\begin{tabular}{ccccccccccc} 
\toprule 
\multicolumn{3}{c}{\textbf{X-Domain}} && \multicolumn{3}{c}{\textbf{FrAC (A)}} && \multicolumn{3}{c}{\textbf{FrAC (C)}}\\ 
\textbf{P}&\textbf{R}&\textbf{F1}&&\textbf{P}&\textbf{R}&\textbf{F1}&&\textbf{P}&\textbf{R}&\textbf{F1}\\
\midrule 
0.65 & 0.59  & 0.59 && 0.74 & 0.81 &  0.77    && 0.84 & 0.85 & 0.83     \\
\bottomrule
\end{tabular}%
\caption{Precision, Recall and Macro-F1 scores of RoBERTa cross-domain (X-Domain) and on FrAC articles (A) and comments (C).}
\label{tab:f1_roberta_frac}
\end{table}


\paragraph{Evaluation}

\begin{table*}[t]
\centering
\small
\begin{tabular}{lrrrcrrr}
\toprule
\multirow{2}{*}{\textbf{Topic}} & \multicolumn{3}{c}{\textbf{SOCC}} && \multicolumn{3}{c}{\textbf{NYT}} \\
\cline{2-4} \cline{6-8}
& \textbf{Articles} & \textbf{Comments} & \textbf{Avg. C/A} && \textbf{Articles} & \textbf{Comments} & \textbf{Avg. C/A} \\
\midrule
Gun control      & 20  & 855   & 42  && 112 & 28,137 & 251 \\
Russia-Ukraine         & 77  & 2,171 & 28  && 310 & 100,355 & 323 \\
Trump \& Elections           & 180 & 7,780 & 43  && 247 & 61,550  & 249 \\
Healthcare       & 53  & 1,636 & 30  && 326 & 67,124  & 205 \\
Immigration      & 115 & 4,394 & 38  && 172 & 33,060  & 192 \\
LGBT+ Rights       & 23  & 803   & 34  && 22  & 3,279   & 149 \\
Education        & 90  & 3,907 & 43  && 78  & 11,614  & 149 \\
Abortion         & 17  & 508   & 29  && 38  & 6,193   & 163 \\
Israel-Palestine         & 53  & 1,794 & 34  && 63  & 7,417   & 118 \\
Climate Change   & 311 & 11,271 & 36 && 231 & 23,990  & 104 \\
Syria \& IS           & 138 & 4,195 & 30  && 72  & 3,757   & 52 \\
\midrule
\textbf{Total}   & \textbf{1,077} & \textbf{39,614} & \textbf{36.8} && \textbf{1,699} & \textbf{346,476} & \textbf{204} \\
\bottomrule
\end{tabular}
\caption{SOCC and NYT data statistics by topic and source, including the average number of comments per article, after the pre-processing explained in Section~\ref{sec:methodology}.}
\label{tab:dataset_stats}
\end{table*}

The averaged results in Table~\ref{tab:f1_avg} show little performance difference across models, with \texttt{RoBERTa-Large} slightly outperforming the larger \texttt{Qwen2.5-Instruct} and \texttt{LLaMA3.1-8B-Instruct}. This result aligns with recent literature on frame classification, where BERT-based pre-trained models have consistently demonstrated strong performance~\citep{sajwani2024frappe, daffara2025generalizabilitymediaframescorpus}. Since RoBERTa also requires substantially less fine-tuning time (approximately 20 minutes versus 13 hours for PEFT-tuning of LlaMA and Qwen), all remaining experiments in this paper are based on fine-tuned RoBERTa. Table \ref{tab:f1_all} reports the per-label breakdown for the best performing model, RoBERTa. 

To evaluate model generalizability to unseen topics, we train RoBERTa on all but one MFC topics, evaluate on the held-out topic, average over all five combinations. Table~\ref{tab:f1_roberta_frac} (X-Domain) revealed an expected, albeit relatively small drop in RoBERTa performance (from 0.66 In-Domain in \Cref{tab:f1_all} to 0.59), suggesting a reasonable ability to generalize to unseen topics, which is important because the model will be applied to topics not present in the MFC training set.

Finally, we evaluated RoBERTa on the annotated FrAC corpus (Table~\ref{tab:f1_roberta_frac}, FrAC), i.e., a data set that differs from the MFC fine-tuning data in its outlets, time points and distinct yet overlapping set of topics (\Cref{sec:frac} and \Cref{tab:dataset_stats}). The model achieved strong performance on FrAC with an F1 score of 0.77 on articles and 0.83 on comments, demonstrating effective generalization to reader comments despite being trained primarily on formal news articles.\footnote{FrAC was annotated directly at the sentence level using the consolidated 9+1 frame taxonomy, whereas MFC labels were created post-hoc from span-level annotations, introducing label noise which likely negatively affected model performance.} This result validates our central methodological contribution: that sentence-level frame classification can successfully bridge the gap between formal news articles and informal reader comments. 

Having validated RoBERTa's performance on our gold-standard FrAC corpus, we then applied it to the full sentence-tokenized SOCC and NYT data sets (\Cref{ssec:data}) for our large-scale framing effects analysis.\footnote{We used the NLTK library (\url{https://www.nltk.org/}) for sentence tokenization of article and comment texts.}

\subsection{Dominant Frame Detection}
\label{sec:dominant}
We analyze framing effects between articles and comments in terms of their respective dominant frames as well as article secondary frames. We define the dominant frame as the most frequent sentence-level frame label in a given document. Secondary frames are all other frames that were assigned to at least one sentence.
To reduce noise from near ties, we only assign a dominant frame if the most frequent frame meets both of the following criteria: it appears in at least 3 sentences and it represents at least 40\% of all sentences in the document. We make an exception for single-sentence texts (comments), whose frame is automatically considered to be dominant. Articles and comments whose most frequent frame fails to meet these combined thresholds were excluded from further analysis.

\paragraph{Evaluation} To assess the reliability of our dominant frame detection heuristics, two authors of the paper examined a sample of 110 articles and comments (55 for each) and assessed if the dominant frame derived by applying the heuristics explained above is correct. The agreement rate between the annotators was 83\%. Furthermore, human annotations agreed with the heuristic prediction 88\% of the time, showing that our approach effectively captures the dominant frame for both articles (95\%) and comments (82\%).

\subsection{Topic Extraction}
\label{sec:topic}
We apply BERTopic \citep{grootendorst2022bertopicneuraltopicmodeling} separately to SOCC and NYT to extract topics from news articles, assigning each article a probability distribution across all identified topics.\footnote{We report the full list of parameters used in Appendix~\ref{appendix:bertopic}.}
We analyzed the initial hierarchical topic structures produced by BERTopic and manually merged closely positioned topics with small inter-topic distances and high thematic similarity. The refined topic model was then applied to reader comments. We ultimately assigned each article and each comment the single most likely topic predicted by BERTopic.

This process was conducted independently for both corpora, to avoid any bias in subsequent analyses. The extracted topics showed substantial thematic overlap across data sets, enabling cross-dataset comparison. We manually identified eleven overarching, semantically cohesive topics which we use in our downstream analysis (\Cref{tab:dataset_stats}).

\paragraph{Evaluation} To assess the quality of topic assignments, we randomly sampled 5 articles and 5 comments from each of the 11 topic groups, yielding a total of 110 texts. Two authors of the paper independently reviewed each assigned topic and judged whether it was acceptable or not. We observe a high degree of inter-annotator agreement (96\%). The overall accuracy of the model's predicted topics, calculated against the labels assigned by the annotators was found to be 93\% (on both comments and articles).

\subsection{Resulting Data Set}
\label{sec:result_dataset}

With our framework and validated implementation at hand, we next apply it to analyze framing effects across the SOCC and NYT data sets (\Cref{ssec:data}) and eleven selected topics (\Cref{sec:topic}). We only include article-comment pairs where both share the same predicted topic (excluding off-topic comments to study framing effects when discussions remain on-topic) and where dominant frames could be extracted for both articles and comments. We also filter out comments shorter than 5 words, and duplicate comments. This final data set consists of nearly 3,000 articles and over 380,000 associated comments, as detailed in Table~\ref{tab:dataset_stats}.

\section{Framing Effects across Topics and Outlets}
\label{sec:analysis}
We now address our three research questions (Section~\ref{sec:rqs}) asking to what extent dominant article frames stick with readers? How do these trends vary by topic? And, if the dominant frame does not stick, how readers reconstruct the media frame in their comments?


\begin{figure}
\includegraphics[width=\columnwidth]{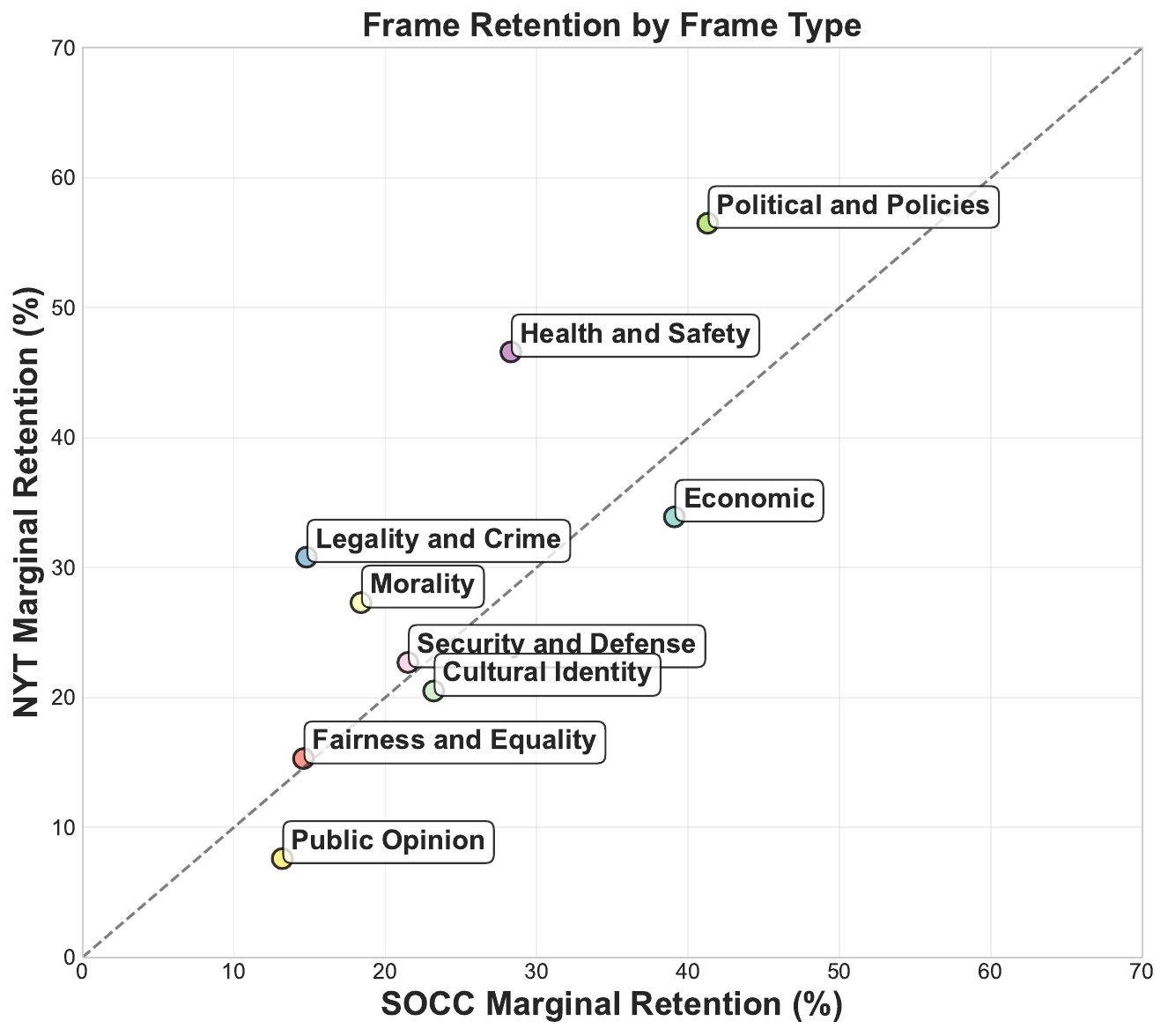}
\caption{Predicted frame retention rates in reader comments based on regression-derived marginal effects. X- and Y-axes show predicted retention for SOCC and NYT, respectively. Retention varies by frame type, and the predictions correlate strongly across the two media outlets.}

\label{fig:retention_type}
\end{figure}

\subsection{To what extent are dominant article frames retained in reader comments?}
\label{sec:frame_type}
In line with our definition of frame retainment as the echoing of the {\it dominant} frame of a news article, we first investigate how often the dominant frame kept in its associated direct readers comments. We do so across the New York Times (NYT; USA) and The Globe and Mail (SOCC; Canada).

Overall, we find that NYT readers demonstrate stronger frame retention (50.7\%) compared to SOCC readers (37.3\%). Conversely, at least half of the comments (almost two thirds in SOCC) {\it change} away from the dominant article frame indicating that readers often do not retain the original primary framing of the article. In section~\ref{ssec:reframe}, we analyze these reframing patterns in depth.

We next investigate how this pattern varies by frame type. We do so through a logistic regression mixed effects model that predicts frame retention (i.e., comment frame being identical to the dominant article frame) as a function of fixed effects for dominant article frame and article topic, and a random effect controlling for article ID,\footnote{We control for article ID due to repeated measurements, given that we have multiple comments per article.}
\begin{align}
\label{equation}
    y_i \sim \text{topic}_d + \text{frame}_d + (1|\text{id}_d), 
\end{align}
where the input is a document-response pair $i=(d,r)$, $y_i$ is a binary indicator of frame retention ($\text{frame}_d{==}\text{frame}_r$). 
\begin{figure}
    \centering
    \includegraphics[width=\linewidth]{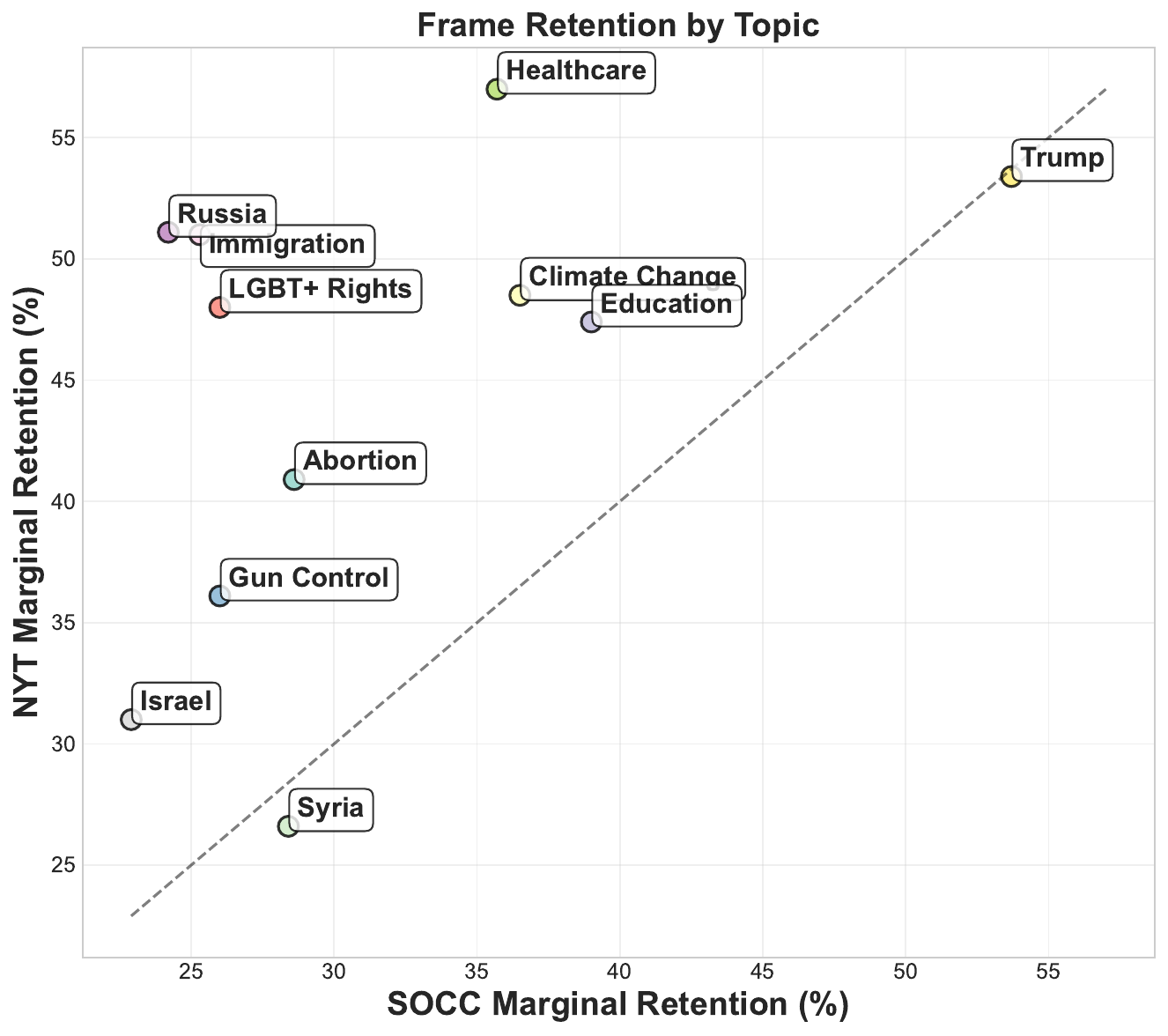}
    \caption{Predicted topic retention rates in reader comments based on regression-derived marginal effects. Points above the diagonal line indicate higher predicted retention in NYT, while points below indicate higher predicted retention in SOCC. Topic names are shortened for readability.}
    \label{fig:topiceffects}
\end{figure}

Here, we are interested in the marginal effect of $\text{frame}_d$ on $y$ as derived from the fitted model coefficients. 
Figure~\ref{fig:retention_type} displays the frame retention patterns separately for each dominant article frame. Looking at the spread of frames along the x- and y-axis independently, we can see that the extent of retention varies by frame. Specifically, frames such as ``Political and Policies'', ``Economic'', and ``Health and Safety'' show high retention rates. In contrast, ``Public Opinion'', ``Fairness and Equality'', and ``Cultural Identity'' exhibit substantially lower retention rates. This suggests that readers are more likely to reframe content that is presented through frames that require alignment with personal beliefs or values to be accepted \citep{chong2007framing}.

To formally test whether the frame choices in reader comments were influenced by the frames used in the corresponding articles -- that is, whether the distribution of frames in comments is not independent of the article's dominant frame -- we conducted chi-squared tests of independence. For each source frame $f$, we tested the null hypothesis that the presence of $f$ in an article is independent of the presence of $f$ in a comment. For all frames, we were able to reject the null hypothesis of independence ($p \ll .001$), although with small to moderate effect sizes (e.g., small effects: ``Public Opinion'' Cramér's V = 0.065 in NYT; moderate effects: ``Economic'' V = 0.357 in SOCC). 

Figure~\ref{fig:retention_type} additionally reveals that the reframing patterns are remarkably consistent across NYT and SOCC, i.e., all points cluster close to the diagonal line of perfect correlation. This cross-outlet consistency addresses a key limitation of previous single-outlet studies and suggests that our uncovered patterns might reveal a general trend in news reader behavior, rather than being idiosyncratic to a particular outlet or sample.

\subsection{How does frame retention vary across topics?}
While previous studies have exclusively focused on a single topic and have often produced contradictory results, here we can more rigorously inspect the variation of framing effects across topics under our unified framework.

We explore the effect of topics on framing effects using the same regression framework as in Section~\ref{sec:frame_type}, but this time focusing on the marginal effects of topic on retention from equation~(\ref{equation}). This provides predicted frame retention probabilities per topic, controlling for frame type and article-specific effects. The resulting predicted probabilities of frame retention per topic are shown in Figure~\ref{fig:topiceffects} for NYT (y-axis) and SOCC (x-axis).

We find that predicted framing effects vary considerably by topic, with notable differences between outlets. Several topics show consistent retention patterns across both outlets. Trump \& Elections, for instance, demonstrates high predicted retention in both outlets. In contrast, international conflicts like Israel-Palestine and Syria \& IS show consistently low retention across both outlets, clustering in the lower portion of the distribution.

Other topics reveal substantial outlet-specific differences. Immigration and Healthcare show notably higher predicted retention in NYT than SOCC, as does coverage of the Russia-Ukraine conflict and LGBT+ Rights, as shown in Figure~\ref{fig:topiceffects}.

These outlet-specific variations, which have not been examined in previous works, confirm that "different kinds of issues are interpreted by the media and by the public in different ways" \citep[pg.~17]{neuman1992common}, suggesting that topic-based retention and reframing patterns are more outlet-specific than the frame-based patterns in Figure~\ref{fig:retention_type}. This finding is in line with the inconsistent results reported in previous small-scale, single-topic studies in the social sciences (Sections~\ref{sec:related_work} and \ref{sec:rqs}). 

In this paper, we deliberately remain on the {\it descriptive} level and refrain from {\it explaining} the observed differences, being cogniscent of many confounding factors that may be at play. However, our framework paves the way for future causal analyses of framing across different topics and news outlets.

\subsection{If readers reject the dominant article frame, how do they reframe?}
\label{ssec:reframe}
Prior communication studies have used varied approaches to analyze reframing - from binary annotations of the presence of frames (e.g.,\citet{holton_commenting_2014, muniz_shaping_2015}) to dominant frame identification (e.g., \citet{zhou_parsing_2007, koteyko_climate_2013}). However, none systematically examined \textit{how} readers reframe when they diverge from the dominant article frame -- whether they introduce entirely novel frames or selectively emphasize non-dominant frames already present in the source. Here, we address this question directly.

Comments change away from the dominant article frame 50\% (NYT) or 63\% (SOCC) of the time, so a natural question is {\it how} readers reframe articles. In our analysis, we distinguish between two types of reframing: (a) \textbf{complete reframing}, where the comment frame does not appear in the source article at all, and (b) \textbf{selective reframing}, where comments adopt secondary frames which are present in the source article, but do not dominate the message. 
In SOCC, 11.7\% of comment frames exhibit complete reframing, while 62.8\% show selective reframing. In NYT, these fractions are 5.2\% and 49.3\%. respectively. This shows that, rather than completely accepting or rejecting the dominant source framing, audiences predominantly engage in selective interpretation by adopting secondary frames from the article.

Figure~\ref{fig:reframing_nyt} shows reframing patterns for the NYT. Focusing first on complete reframing (\Cref{fig:nyt_complete}), Political framing dominates source articles but commentators introduce diverse alternative frames. In particular readers reframe in terms of value-laden frames including ``Morality'' (increasing from 0.8\% in articles to 26.4\% in Comments) or ``Cultural Identity'', as well as ``Economic'' framing. This pattern holds across both NYT and SOCC (as shown in \Cref{app:reframing_socc_NEW}).

Selective reframing patterns in NYT are shown in Figure~\ref{fig:nyt_selective}. Here, the frame distributions in the comments align more closely with the dominant article frame distributions, although we again observe an increase in value-laden frames (``Morality'', ``Health and Safety'') as well as ``Economic'' framing.

\paragraph{Topic-specific Reframing}
We investigate topic-specific reframing patterns for topics that included at least 100 articles. For instance, in the case of \textit{Immigration} (Table~\ref{tab:immigration_reframing}, left), SOCC readers engage in selective reframing by emphasizing both ``Cultural Identity'' frames when ``Political and Policies'' dominate the article, and vice versa, suggesting a tight interconnection of these two perspectives. NYT readers show different selective reframing patterns, primarily shifting from ``Legality and Crime'' frames to ``Political and Policies'' frames, and from ``Political and Policies'' to ``Economic frames'', indicating a tendency toward more policy-oriented and economic interpretations of immigration coverage. Complete reframing differs substantially (Table~\ref{tab:immigration_reframing}, right). In NYT, complete reframing shows ``Economic'' frames being introduced when they were entirely absent from articles. Again, both outlets reframe more objective (economic, political, legal) views into value-laden frames (cultural, morality, fairness). Tables~\ref{tab:selective_reframing} and \ref{tab:complete_reframing} in the Appendix show selective and complete reframing analysis for additional topics, revealing substantial topic-specific differences.

In conclusion, the results discussed here align with the constructionist view of framing as "interactive, vulnerable and prone to counter-frames" \citep[pg.~76]{vangorp2007constructionist}, with audiences engaging in topic-specific patterns of selective retention for frames that resonate with them, whilst actively reframing those that do not.

\begin{figure*}[t]
    \centering
    \begin{subfigure}[t]{0.49\linewidth}
        \centering
        \includegraphics[clip,trim=1cm 1.2cm 1cm 1cm, width=\linewidth]{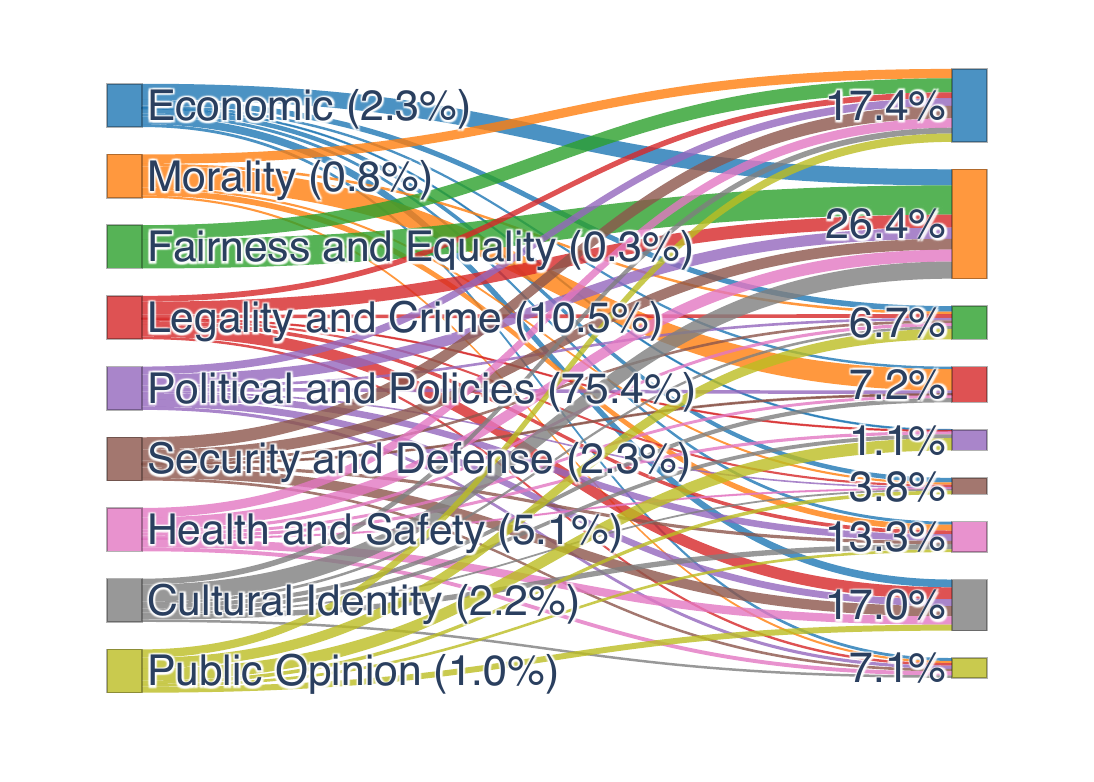}
        \caption{Complete reframing.} 
        \label{fig:nyt_complete}
    \end{subfigure}
    \hfill
    \centering
    \begin{subfigure}[t]{0.49\linewidth}
        \centering
        \includegraphics[clip,trim=1cm 1cm 1cm 1cm,width=\linewidth]{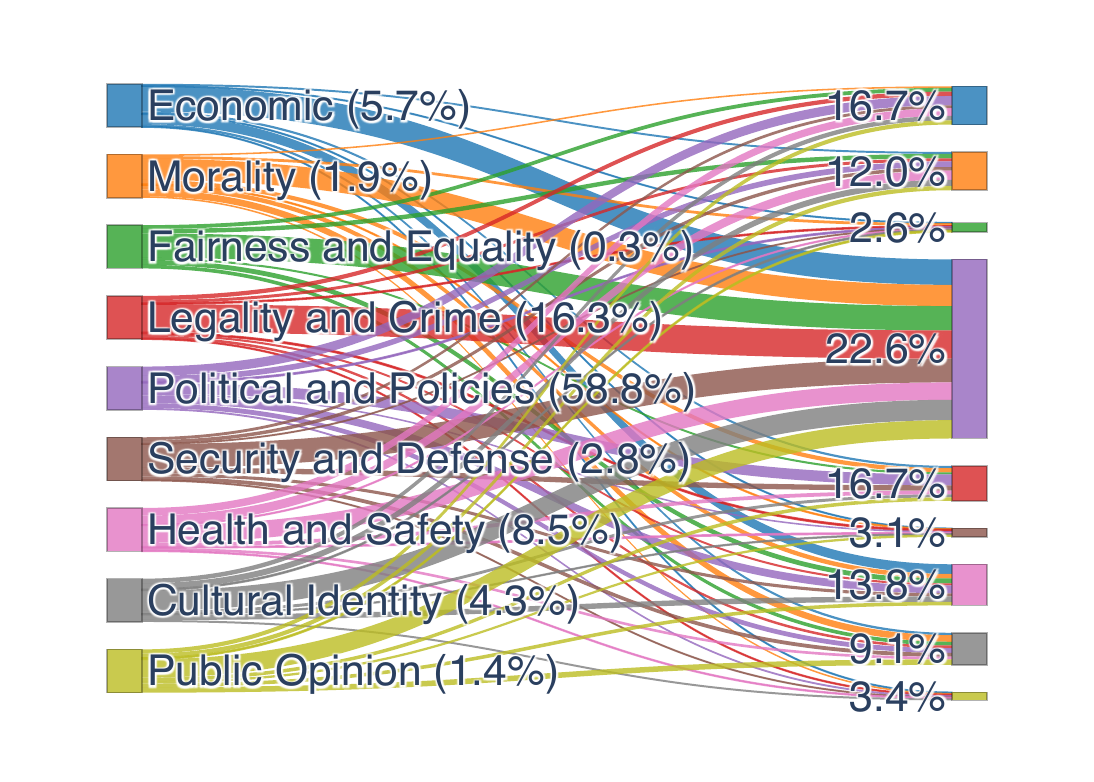}
        \caption{Selective reframing.} 
        \label{fig:nyt_selective}
    \end{subfigure}
    \caption{Reframing in the New York Times. (a) Complete reframing patterns where readers use frames that never appear anywhere in the original articles. (b) Selective reframing where readers emphasize secondary frames that differ from the dominant article frame but exist elsewhere in the article. Left percentages show dominant article frame distributions; right percentages show comment frame distributions among complete (a) / selective (b) reframing cases.}
    \label{fig:reframing_nyt}
\end{figure*}


\begin{table}[t]
\centering
\footnotesize
\begin{tabular}{lll}
\toprule
&{Selective Reframing} & {Complete Reframing } \\
\midrule
 \multirow{5}{*}{\rotatebox{90}{SOCC}}
& Cultural  $\rightarrow$ Political  & Political  $\rightarrow$ Cultural  \\
& Health $\rightarrow$ Cultural  & Economic $\rightarrow$ Political  \\
& Political  $\rightarrow$ Cultural  & Health  $\rightarrow$ Morality \\
& Health $\rightarrow$ Political  & Economic $\rightarrow$ Cultural  \\
& Economic $\rightarrow$ Political  & Political  $\rightarrow$ Fairness \\
\midrule
 \multirow{5}{*}{\rotatebox{90}{NYT}}
& Legality  $\rightarrow$ Political  & Political  $\rightarrow$ Economic \\
& Legality  $\rightarrow$ Economic & Political  $\rightarrow$ Legality  \\
& Political  $\rightarrow$ Economic & Legality  $\rightarrow$ Cultural  \\
& Political  $\rightarrow$ Legality  & Political  $\rightarrow$ Cultural  \\
& Political  $\rightarrow$ Cultural  & Political  $\rightarrow$ Morality \\
\bottomrule
\end{tabular}
\caption{Reframing ``Immigration'' in SOCC (top) and NYT (bottom). Each cell shows the five most common selective (left) and complete (right) shifts from dominant article $\rightarrow$ comment frame. Frame names are abbreviated to their first word.}
\label{tab:immigration_reframing}
\end{table}

\section{Discussion}

Unlike previous computational approaches to framing detection and analysis, our study is centered on a \textit{constructionist} understanding of framing as an interactive process where readers selectively engage, modify, and reconstruct media frames based on their own interpretative resources ~\citep{neuman1992common, scheufele1999framing, vangorp2007constructionist}. While existing computational methods treat framing as isolated classification tasks focused solely on source content, we examined the dynamic interaction between media framing and audience reception. In uncovering systematicities (RQ1; \Cref{fig:retention_type}) and topic-specific differences (RQs 2-3) in a large-scale study, we bridged the disconnect between large-scale yet isolated NLP approaches and small-scale communication studies that lack generalizability.

In applying one unified framing conceptualization across eleven topics, our findings suggest that inconsistent results on reframing patterns in prior work can be partially explained by topic- and outlet-specific variation in framing effects. We also identified consistent trends, such as value-laden frames (``Fairness and Equality'', ``Morality'') being more likely to be reframed (Figure~\ref{fig:retention_type}), but also more likely to be introduced in the context of complete reframing (Figure~\ref{fig:nyt_complete}). This divergent pattern suggests that audiences reject value-laden framings that conflict with their worldview while actively introducing those that align with it. This extends theoretical formulations about audience personal formulation of messages \citep{chong2007framing} by showing that readers not only selectively accept frames matching their values, but also actively impose their preferred value frames onto discussions where they were originally absent.

In distinguishing selective from complete reframing, we show that readers predominantly reframe in selective ways, providing empirical evidence for what framing theory has long posited: that news consumers reformulate and engage with content selectively based on their preferences \citep{neuman1992common, scheufele1999framing} -- a phenomenon that small-scale social science studies could not disentangle systematically, and that NLP studies have overlooked by disregarding response analysis.

Our analysis reveals that frame-based patterns are relatively stable across outlets, while topic-specific effects vary more widely. These differences likely arise from complex interactions between unobservable factors including reader demographics and national context. Furthermore, our data sets conflate country (US versus Canada), editorial leaning (center-left NYT versus center-right SOCC), and time period (2012–2016 NYT versus 2017–2018 SOCC). We hence deliberately maintain a {\it descriptive} level and leave disentanglement of these factors for future work.

A key methodological contribution lies in demonstrating reliable frame classification across fundamentally different text genres -- formal news articles and informal reader comments. This cross-genre generalization addresses a critical gap in computational framing analysis, where models typically operate within homogeneous text types. Our validation shows strong performance on our human-annotated data set FrAC.

Our study builds on the MFC, a widely-used framework in NLP. Yet, it is by no means the only and optimal frameworks, and it has been criticized for the broad nature and lack of theoricatical ground of its framings~\citep{ali_survey_2022}. We note that our method is framework-agnostic: alternative framing theories could be readily integrated, whether more generic frameworks (e.g., \citet{semetko2000framing}), or issue-specific taxonomies tailored to particular domains. Similarly, our approach supports both large-scale analyses of framing trends and more fine-grained investigations that future work could further explore the role of minority frames.

Finally, our framework holds potential relevance for other NLP tasks including argumentation interpretation and generation, and disinformation detection. The systematic patterns we identify in how audiences reconstruct frames could inform models of persuasive text generation, helping systems understand how different framings might be received and transformed by target audiences.

\section{Conclusion}

This work addresses the gap between framing theory and its computational operationalization. While framing theory posits that individuals respond differentially to framed information -- accepting, rejecting, or modifying frames based on their cognitive schemas -- computational approaches have largely reduced framing to taxonomic classification, neglecting audience reception processes.

We introduced a scalable computational framework for analyzing media framing effects as an interactive process between source content and audience responses. Through analysis of over 3,000 news articles and 380,000 comments across two outlets and eleven topics, we demonstrate that readers reframe content at least as frequently as they retain original frames, often reframing selectively by picking up on secondary article frames. Frame retention varies systematically by frame type with objective frames (e.g., ``Health \& Safety'') more likely to be retained than value-laden frames (e.g., ``Morality''). We contributed a robust framing scheme with classifiers that work across document types, providing a foundation for future research. 

Our analysis leaves open questions for future work, including how framing effects propagate through nested comment conversations and identifying causal linguistic properties that explain specific framing effects, or thread and temporal analysis.
    
\section*{Acknowledgment}
This paper was written with the support from the Melbourne Research Scholarship by the University of Melbourne provided to MG. This work was supported by the Australian Research Council Discovery Early Career Research Award (Grant No. DE230100761). We thank Damian Curran, Thomas Hollow, Jemima Kang, Tom Monaghan, Gordon William Smith, and Katie Warburton for their annotations.

\bibliography{new}
\bibliographystyle{acl_natbib}

\appendix

\iftaclpubformat
\onecolumn
\fi

\setlength{\parskip}{0pt}
\setlength{\itemsep}{0pt}
\setlength{\parsep}{0pt}
\setlength{\topsep}{0pt}

\section{Frame Classifiers}
\label{app:fine-tune}
We report the parameters used for fine-tuning the two LLMs — Llama and Qwen — and RoBERTa for the frame classification task. The LLMs were fine-tuned using Unsloth\footnote{\url{https://unsloth.ai/}}, applying parameter-efficient fine-tuning via Low-Rank Adaptation (LoRA) \citep{hu2022lora}, while RoBERTa was fully fine-tuned. All training was conducted on high-performance computing infrastructure with NVIDIA A100 GPUs. We also include the final prompt used for the task in Table~\ref{tab:prompt-template}.

\vspace{-0.5em}
\subsection{LLaMA3-8B-Instruct and Qwen2.5-Instruct}
\label{qwen-llama-param}
We used the same set of parameters on Unsloth for both models.

\begin{compactitem}
    \item \texttt{Max Sequence Length:} 3000 tokens
    \item \texttt{Batch Size:} 8 (gradient accumulation = 8, effective batch = 64)
    \item \texttt{Number of Epochs:} 6
    \item \texttt{Learning Rate:} 2e-4
    \item \texttt{LoRA Configuration:} r = 16, $\alpha$ = 16, dropout = 0, target modules = \texttt{[q\_proj, k\_proj, v\_proj, o\_proj, gate\_proj, up\_proj, down\_proj]}
    \item \texttt{Precision:} bf16 (if supported) or fp16
    \item \texttt{Optimizer:} AdamW (8-bit)
    \item \texttt{Scheduler:} Linear with 10\% warm-up ratio
    \item \texttt{Regularization:} Weight decay = 0.01
    \item \texttt{Early Stopping:} Patience = 3 epochs
    \item \texttt{Checkpointing:} Save best model by validation loss at end of each epoch
    \item \texttt{Packing:} Enabled 
\end{compactitem}

\vspace{-0.5em}
\subsection{RoBERTa}
\label{roberta-param}

\begin{compactitem}
  \item \texttt{Maximum sequence length:} 70 tokens
  \item \texttt{Data splitting:}
    \begin{itemize}[nosep]
      \item Training set: 80\%
      \item Validation set: 10\%
      \item Test set: 10\%
    \end{itemize}
  \item \texttt{Batch size:} 64
  \item \texttt{Learning rate:} 5e-6
  \item \texttt{Number of epochs:} 10
  \item \texttt{Weight decay:} 0.01
  \item \texttt{Optimizer scheduler:} Linear learning rate scheduler with 5\% warmup
  \item \texttt{Early stopping:} Patience of 2 epochs based on validation F1 score
  \item \texttt{Evaluation and model saving:} After every epoch, best model saved based on F1 score
  \item \texttt{Loss function:} Cross-entropy
\end{compactitem}

\vspace{-0.5em}
\section{LLMs Prompt}
\label{app:prompt}
We experimented with various prompting strategies to balance performance and prompt length. We found that including an elaborated description of the category definitions from \citet{card_media_2015}, along with clear task instructions, consistently yielded the best results. The final prompt is reported in Table~\ref{tab:prompt-template}.

\begin{table*}[ht]
\centering
\small
\begin{tabular}{|p{0.95\linewidth}|}
\hline
\textbf{Prompt Used for Fine-Tuning} \\
\hline

You are an expert in discourse and framing analysis. You are analyzing how texts are framed in their coverage of topics. Your task is to identify the frames used in the text into one of nine frame categories based on how the issue is presented and which aspects are emphasized. \\
\\
\textbf{Instructions:}
\begin{enumerate}
    \item Read the text carefully.
    \item Identify the frame that shapes the article's narrative. Only ONE frame applies.
    \item Output ONLY the NUMERICAL code (e.g. "2") for those frames first, followed by a new line. Do not include any other text or explanation.
\end{enumerate}

1 - Economic: The costs, benefits, or financial implications of the issue (individual, community, or economy)\\

2 - Morality: Perspectives compelled by religion, ethics, or sense of social responsibility\\

3 - Fairness and Equality: Equality or inequality in how laws and resources are applied or distributed\\

4 - Legality and Crime: Laws, judicial interpretations, crime rates, legal consequences\\

5 - Health and Safety: Healthcare, disease, sanitation, mental health, infrastructure safety\\

6 - Cultural Identity: Social norms, values, customs, and cultural preservation\\

7 - Public Opinion: Public attitudes, polling, consequences of diverging from popular opinion\\

8 - Security and Defense: Protection, national security, border security, social stability\\

9 - Political and Policies: Political discourse, partisanship, lobbying, legislative processes\\

10 - Other: If the sentence is not framed in any of the proposed categories, select this label\\

\hline
\end{tabular}
\caption{Prompt template format used for fine-tuning the models.}
\label{tab:prompt-template}
\end{table*}

\vspace{-0.5em}
\section{BERTopic Configuration}
\label{appendix:bertopic}

We trained a BERTopic model \citep{grootendorst2022bertopicneuraltopicmodeling} for clustering textual topics with the following configuration:

\begin{compactitem}
  \item \textbf{Embedding Model:} \texttt{all-mpnet-base-v2} from the SentenceTransformers library.
  
  \item \textbf{Vectorizer:} \texttt{CountVectorizer} with the following parameters:
  \begin{compactitem}
    \item \texttt{stop\_words}: English stopwords from NLTK
    \item \texttt{min\_df}: 5
    \item \texttt{ngram\_range}: (1, 2)
  \end{compactitem}
  
  \item \textbf{UMAP Dimensionality Reduction:}
  \begin{compactitem}
    \item \texttt{n\_neighbors}: 15
    \item \texttt{n\_components}: 5
    \item \texttt{min\_dist}: 0.0
    \item \texttt{metric}: cosine
    \item \texttt{random\_state}: 42
  \end{compactitem}
  
  \item \textbf{Clustering Model:} \texttt{HDBSCAN} with:
  \begin{compactitem}
    \item \texttt{min\_cluster\_size}: 15
    \item \texttt{min\_samples}: 7
    \item \texttt{metric}: euclidean
    \item \texttt{cluster\_selection\_method}: eom
    \item \texttt{prediction\_data}: True
  \end{compactitem}
  
  \item \textbf{BERTopic Parameters:}
  \begin{compactitem}
    \item \texttt{calculate\_probabilities}: True
    \item \texttt{verbose}: True
    \item \texttt{n\_gram\_range}: (1, 2)
    \item \texttt{min\_topic\_size}: 5
  \end{compactitem}
\end{compactitem}

\vspace{-0.5em}
\section{Processed Training Data} 
\label{app:processed_training}
We report the statistics of the final processed data set, as explained in Section~\ref{sec:mfc_theory_classification}. Table~\ref{tab:frame_distribution} reports the number of final samples per label across topics, from both the MFC and \citet{hartmann_issue_2019}. 

\begin{table*}[ht]
\centering
\begin{small}
\begin{tabular}{lrrrrrr|r}
\hline
\textbf{Frame} & \textbf{Guns} & \textbf{Death Penalty} & \textbf{Immig.} & \textbf{Same-Sex} & \textbf{Tobacco} & \textbf{Hartmann} & \textbf{Total} \\
\hline
Legality and Crime     & 4,942  & 9,215  & 4,111  & 3,306  & 1,376  & 344   & 23,294 \\
Political and Policies & 7,405  & 1,006  & 3,436  & 3,357  & 1,853  & 211   & 17,268 \\
Security and Defense   & 947    & 163    & 770    & 11     & -      & -     & 1,891 \\
Health and Safety      & 938    & 740    & 534    & 158    & 1,409  & -     & 3,779 \\
Economic               & 810    & 156    & 1,294  & 441    & 2,303  & 70    & 5,074 \\
Morality               & 183    & 573    & 147    & 999    & 64     & -     & 1,966 \\
Cultural Identity      & 636    & 432    & 722    & 552    & 459    & -     & 2,801 \\
Public Opinion         & 1,089  & 161    & 470    & 967    & 130    & -     & 2,817 \\
Fairness and Equality  & 60     & 677    & 284    & 681    & 34     & -     & 1,736 \\
Other                   & 275    & 451    & 253    & 1,716  & 305    & -     & 3,000 \\
\hline
Total                  & 17,285 & 13,574 & 12,021 & 12,188 & 7,933  & 625   & 63,626 \\
\hline
\end{tabular}
\end{small}
\caption{Distribution of frame categories across issue topics in the training data used for fine-tuning the frame classification models. The data combines the Media Frames Corpus with additional annotations from \citet{hartmann_issue_2019}.} 
\label{tab:frame_distribution}
\end{table*}

\vspace{-1em}
\section{Annotation Instructions and Interface}
This section includes the instructions which were presented to each annotator on the web interface developed for the task (a screenshot of the interface is reported in Figure~\ref{fig:annotation-interface}).

\begin{figure*}[ht]
    \includegraphics[width=\textwidth]{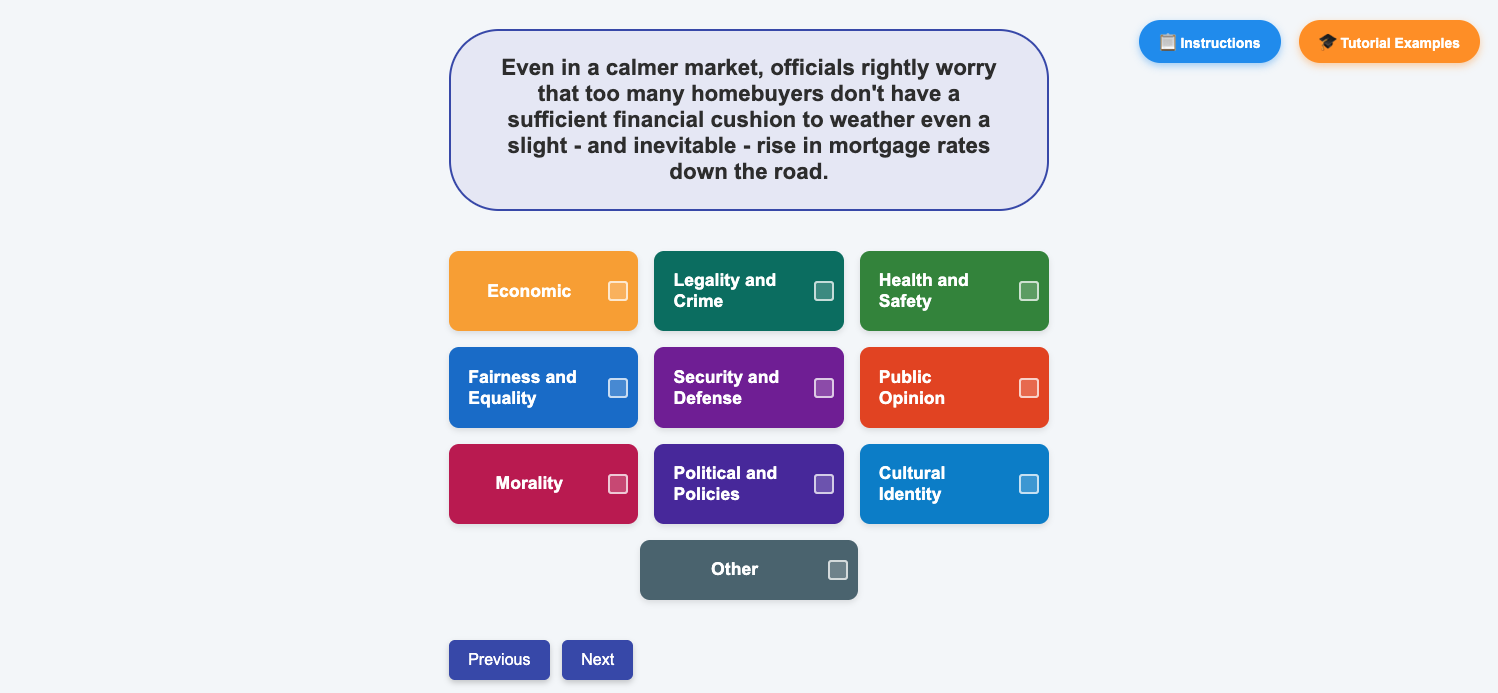}
    \caption{An example from the annotation task on the deployed interface.}
    \label{fig:annotation-interface}
\end{figure*}

\label{annotation-interface}

\vspace{-0.5em}
\subsection*{Understanding the Influence of News Framing on Readers}

Hi there! Thanks for agreeing to participate in this annotation task.

If you're viewing this on a phone or tablet, we recommend switching to a laptop for better comfort and readability.

In this project, we're investigating the different formulations of news in terms of frames.

\vspace{-0.3em}
\subsubsection*{What Are Frames?}

Frames are the perspectives or angles through which information is presented in media. They highlight certain aspects of an issue while downplaying others, influencing how audiences understand and respond to news.

Think of frames as different ``lenses'' that focus attention on specific elements of a story while leaving others in the background.

Frames answer the question: \emph{``How is the issue at hand being presented?''}

\vspace{-0.3em}
\subsubsection*{Your Task}

Your task is to identify which frames are used in short snippets of text from various sources. This helps us understand how different issues are presented and framed in different contexts.

\vspace{-0.3em}
\subsubsection*{How the Annotation Works}

You will see one text at a time. Please read each text carefully and choose the frame that best represents how the information is being conveyed by clicking on the corresponding label (colourful boxes).

If you think more than one frame applies, please select all that you believe are relevant.

If none of the provided categories apply, please select the label ``Other.''

Please note: the examples all focus on same-sex marriage for clarity. The real annotation task covers a variety of different issues.

Once you start annotating, you can click on the ``Instructions'' button at any time to retrieve this information.

\vspace{-0.3em}
\subsubsection*{Framing Categories}
See Table~\ref{tab:framing_categories}
\begin{table*}[h]
\centering
\footnotesize
\begin{tabular}{|p{2.8cm}|p{4.8cm}|p{5.8cm}|}
\hline
\textbf{Frame} & \textbf{Description} & \textbf{Example} \\
\hline
Economic & The costs, benefits, or financial implications of the issue (individual, community, or economy). & \emph{Legalizing same-sex marriage would boost state economies through increased spending on weddings and tourism.} \\
\hline
Morality & Perspectives compelled by religion, ethics, or sense of social responsibility. & \emph{Same-sex marriage is morally wrong and goes against the teachings of our faith.} \\
\hline
Fairness and Equality & Equality or inequality in how laws and resources are applied or distributed. & \emph{Denying same-sex couples the right to marry is a clear violation of equal protection under the law.} \\
\hline
Legality and Crime & Laws, judicial interpretations, crime rates, legal consequences. & \emph{The Supreme Court ruled that bans on same-sex marriage are unconstitutional and violate civil liberties.} \\
\hline
Health and Safety & Healthcare, disease, sanitation, mental health, infrastructure safety. & \emph{Allowing same-sex marriage is an endorsement of unsafe sexual practices and will contribute to the spread of disease.} \\
\hline
Cultural Identity & Social norms, values, customs, and cultural preservation. & \emph{Same-sex marriage threatens the traditional concept of family that our culture has upheld for generations.} \\
\hline
Public Opinion & Public attitudes, polling, consequences of diverging from popular opinion. & \emph{A recent poll shows that 65\% of Americans now support same-sex marriage, up from 30\% a decade ago.} \\
\hline
Security and Defense & Protection, national security, border security, social stability. & \emph{Same-sex marriage could undermine traditional social structures and thus contribute to community stability and security.} \\
\hline
Political and Policies & Political discourse, partisanship, lobbying, legislative processes. & \emph{The governor vetoed a bill legalizing same-sex marriage, citing pressure from conservative lawmakers.} \\
\hline
Other & If the sentence is not framed in any of the proposed categories, select this label. & --- \\
\hline
\end{tabular}
\caption{Framing categories shown on the annotation interface.}
\label{tab:framing_categories}
\end{table*}

\vspace{-0.5em}
\subsubsection*{Practice Phase}

Before starting the actual annotation task, we'll practice together with three examples.

During the practice phase:

\begin{compactitem}
  \item You will be shown \textbf{one sentence at a time}.
  \item You should \textbf{select the appropriate frame label(s)} by clicking on the colored boxes.
  \item After you make your selection, the \textbf{correct answer will appear underneath with an explanation}.
\end{compactitem}

Take your time to read each sentence carefully and consider which perspective or angle is being emphasized.

You can consult the examples you will see during the practice at any time during the annotation by clicking on the ``Tutorial Example'' button on the top-right.

\vspace{-0.3em}
\subsubsection*{Final Notes}

If you have any questions, please contact the organiser.

For more information regarding the framing labels and theory, please refer to the \href{https://minio.la.utexas.edu/compagendas/codebookfiles/Policy_Frames_Codebook.pdf}{Policy Frames Codebook} and this \href{https://aclanthology.org/P15-2072.pdf}{paper}.

Happy annotating!

\vspace{-1em}
\section{Reframing Patterns by Topic}
Figure \ref{tab:selective_reframing} and Figure \ref{tab:complete_reframing} show the 5 most frequent selective and complete reframing patterns in both news outlets in coverage of climate change, immigration, Syria \& IS, Trump \& elections.
\begin{table}[h]
\centering
\scriptsize
\begin{tabular}{l|l|p{4.2cm}|r}
\hline
\textbf{Topic} & \textbf{Outlet} & \textbf{Selective Reframing Pattern} & \textbf{Count} \\
\hline
\multirow{10}{*}{CC} 
& \multirow{5}{*}{SOCC} & Political $\to$ Economic & 1001 \\
& & Economic $\to$ Political & 854 \\
& & Political $\to$ Public Opinion & 113 \\
& & Economic $\to$ Cultural & 90 \\
& & Political $\to$ Cultural & 79 \\
\cline{2-4}
& \multirow{5}{*}{NYT} & Political $\to$ Economic & 1585 \\
& & Economic $\to$ Political & 857 \\
& & Political $\to$ Health & 428 \\
& & Political $\to$ Cultural & 345 \\
& & Political $\to$ Legality & 232 \\
\hline
\multirow{10}{*}{IM} 
& \multirow{5}{*}{SOCC} & Cultural $\to$ Political & 132 \\
& & Health $\to$ Cultural & 113 \\
& & Political $\to$ Cultural & 99 \\
& & Health $\to$ Political & 96 \\
& & Economic $\to$ Political & 85 \\
\cline{2-4}
& \multirow{5}{*}{NYT} & Legality $\to$ Political & 1388 \\
& & Legality $\to$ Economic & 955 \\
& & Political $\to$ Economic & 927 \\
& & Political $\to$ Legality & 903 \\
& & Political $\to$ Cultural & 769 \\
\hline
\multirow{10}{*}{SY} 
& \multirow{5}{*}{SOCC} & Political $\to$ Security & 352 \\
& & Security $\to$ Political & 326 \\
& & Security $\to$ Morality & 78 \\
& & Security $\to$ Cultural & 64 \\
& & Political $\to$ Morality & 41 \\
\cline{2-4}
& \multirow{5}{*}{NYT} & Security $\to$ Political & 321 \\
& & Health $\to$ Political & 146 \\
& & Health $\to$ Morality & 123 \\
& & Security $\to$ Morality & 117 \\
& & Security $\to$ Cultural & 79 \\
\hline
\multirow{10}{*}{TR} 
& \multirow{5}{*}{SOCC} & Political $\to$ Cultural & 319 \\
& & Political $\to$ Public Opinion & 256 \\
& & Political $\to$ Economic & 146 \\
& & Cultural $\to$ Political & 145 \\
& & Political $\to$ Fairness & 108 \\
\cline{2-4}
& \multirow{5}{*}{NYT} & Political $\to$ Cultural & 2136 \\
& & Political $\to$ Morality & 2070 \\
& & Cultural $\to$ Political & 1145 \\
& & Political $\to$ Health & 983 \\
& & Political $\to$ Economic & 774 \\
\hline
\end{tabular}
\caption{Top 5 selective reframing patterns by topic and outlet. CC = Climate Change, IM = Immigration, SY = Syria, TR = Trump.}
\label{tab:selective_reframing}
\end{table}

\begin{table}[ht]
\centering
\scriptsize
\begin{tabular}{l|l|p{4.2cm}|r}
\hline
\textbf{Topic} & \textbf{Outlet} & \textbf{Complete Reframing Pattern} & \textbf{Count} \\
\hline
\multirow{10}{*}{CC} 
& \multirow{5}{*}{SOCC} & Political $\to$ Economic & 125 \\
& & Political $\to$ Cultural & 106 \\
& & Political $\to$ Health & 72 \\
& & Political $\to$ Public Opinion & 68 \\
& & Economic $\to$ Cultural & 66 \\
\cline{2-4}
& \multirow{5}{*}{NYT} & Political $\to$ Morality & 473 \\
& & Political $\to$ Health & 258 \\
& & Political $\to$ Cultural & 120 \\
& & Health $\to$ Morality & 62 \\
& & Economic $\to$ Morality & 51 \\
\hline
\multirow{10}{*}{IM} 
& \multirow{5}{*}{SOCC} & Political $\to$ Cultural & 32 \\
& & Economic $\to$ Political & 23 \\
& & Health $\to$ Morality & 22 \\
& & Economic $\to$ Cultural & 17 \\
& & Political $\to$ Fairness & 17 \\
\cline{2-4}
& \multirow{5}{*}{NYT} & Political $\to$ Economic & 384 \\
& & Political $\to$ Legality & 236 \\
& & Legality $\to$ Cultural & 189 \\
& & Political $\to$ Cultural & 181 \\
& & Political $\to$ Morality & 177 \\
\hline
\multirow{10}{*}{SY} 
& \multirow{5}{*}{SOCC} & Security $\to$ Political & 79 \\
& & Security $\to$ Cultural & 56 \\
& & Political $\to$ Morality & 55 \\
& & Security $\to$ Morality & 49 \\
& & Political $\to$ Cultural & 39 \\
\cline{2-4}
& \multirow{5}{*}{NYT} & Security $\to$ Morality & 48 \\
& & Security $\to$ Cultural & 27 \\
& & Security $\to$ Economic & 25 \\
& & Fairness $\to$ Morality & 14 \\
& & Health $\to$ Cultural & 14 \\
\hline
\multirow{10}{*}{TR} 
& \multirow{5}{*}{SOCC} & Political $\to$ Cultural & 88 \\
& & Political $\to$ Fairness & 70 \\
& & Political $\to$ Economic & 60 \\
& & Political $\to$ Legality & 60 \\
& & Political $\to$ Morality & 38 \\
\cline{2-4}
& \multirow{5}{*}{NYT} & Political $\to$ Morality & 446 \\
& & Political $\to$ Health & 379 \\
& & Political $\to$ Economic & 265 \\
& & Political $\to$ Legality & 252 \\
& & Political $\to$ Cultural & 237 \\
\hline
\end{tabular}
\caption{Top 5 complete reframing patterns by topic and outlet. CC = Climate Change, IM = Immigration, SY = Syria, TR = Trump.}
\label{tab:complete_reframing}
\end{table}

\vspace{-0.8em}
\section{Reframing in SOCC}
Figure~\ref{fig:reframing_socc_complete} and Figure~\ref{fig:reframing_socc_selective} show complete and selective reframing patterns in SOCC, for comparison against NYT results.
\label{app:reframing_socc_NEW}

\begin{figure}[htb!]
    \centering
    \includegraphics[width=0.95\linewidth]{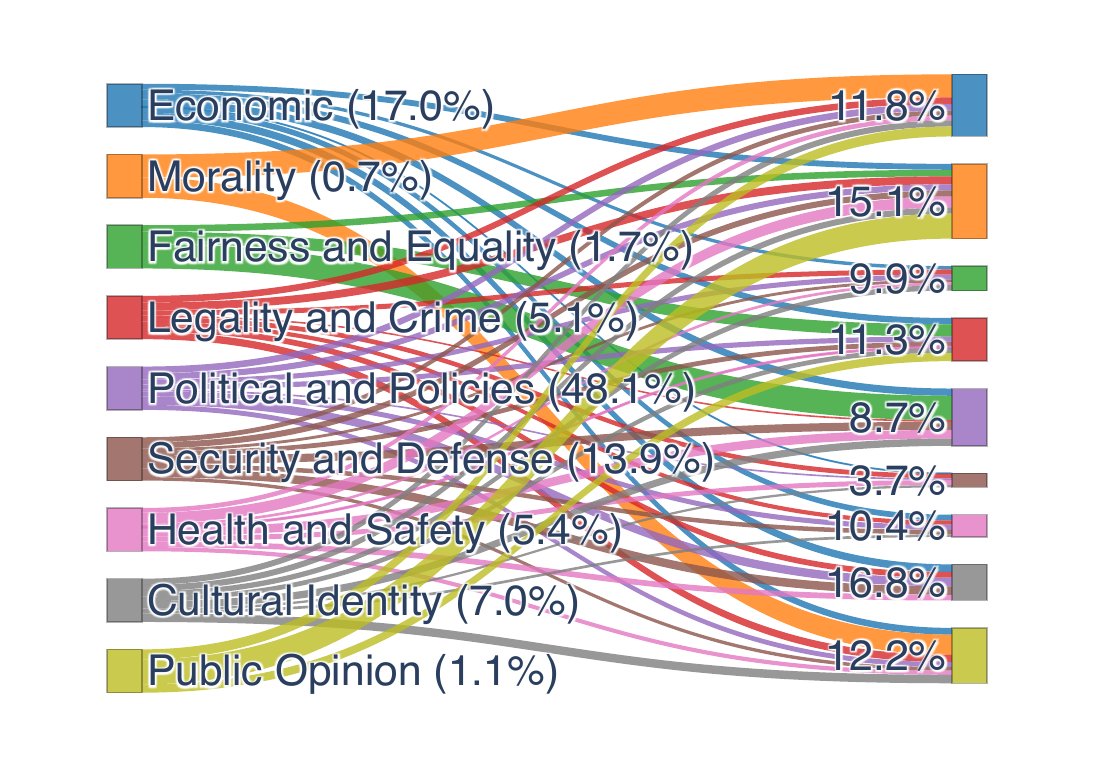}
    \caption{Complete reframing in SOCC showing cases where readers use frames never appearing in articles.}
    \label{fig:reframing_socc_complete}
\end{figure}

\begin{figure}[htb!]
    \centering
    \includegraphics[width=0.95\linewidth]{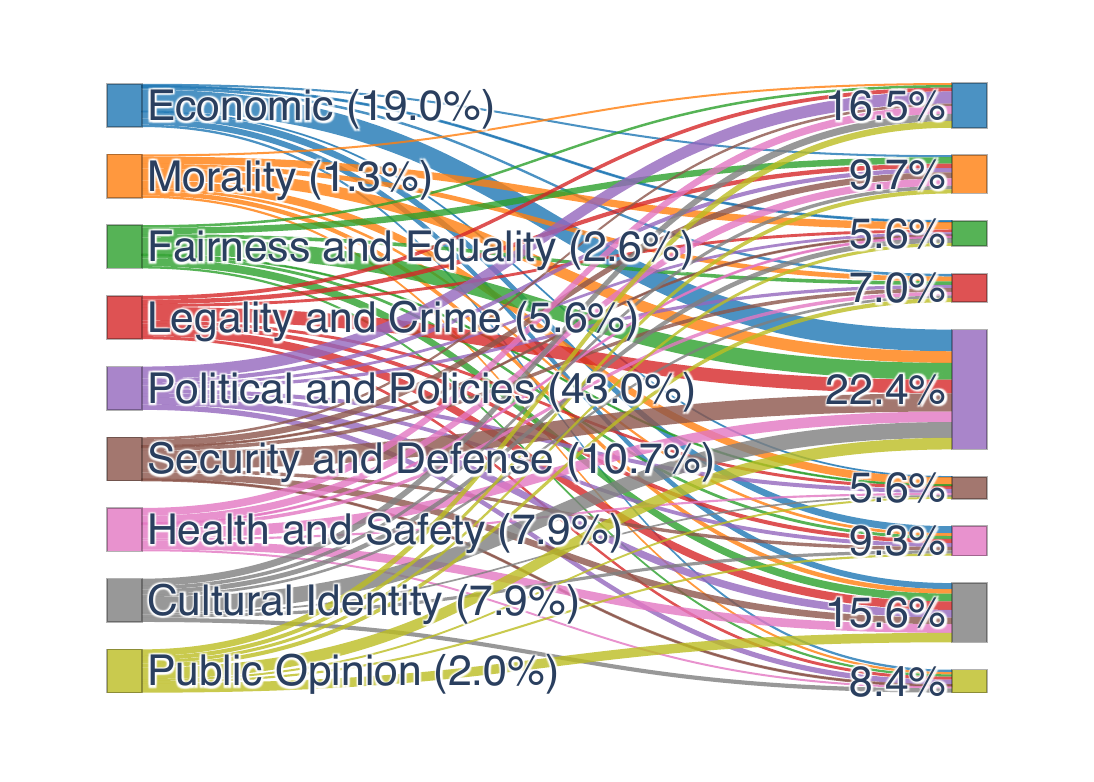}
    \caption{Selective reframing in SOCC showing cases where readers emphasize secondary frames differing from dominant article frames but existing elsewhere in articles.}
    \label{fig:reframing_socc_selective}
\end{figure}

\end{document}